# Best-First and Depth-First Minimax Search in Practice

Aske Plaat, Erasmus University, *plaat@theory.lcs.mit.edu*
Jonathan Schaeffer, University of Alberta, *jonathan@cs.ualberta.ca*
Wim Pijls, Erasmus University, *whlmp@cs.few.eur.nl*
Arie de Bruin, Erasmus University, *arie@cs.few.eur.nl*

Erasmus University,
Department of Computer Science,
Room H4-31, P.O. Box 1738,
3000 DR Rotterdam,
The Netherlands

University of Alberta,
Department of Computing Science,
615 General Services Building,
Edmonton, Alberta,
Canada T6G 2H1

**Abstract**

Most practitioners use a variant of the Alpha-Beta algorithm, a simple depth-first procedure, for searching minimax trees. SSS*, with its best-first search strategy, reportedly offers the potential for more efficient search. However, the complex formulation of the algorithm and its alleged excessive memory requirements preclude its use in practice. For two decades, the search efficiency of "smart" best-first SSS* has cast doubt on the effectiveness of "dumb" depth-first Alpha-Beta.

This paper presents a simple framework for calling Alpha-Beta that allows us to create a variety of algorithms, including SSS* and DUAL*. In effect, we formulate a best-first algorithm using depth-first search. Expressed in this framework SSS* is just a special case of Alpha-Beta, solving all of the perceived drawbacks of the algorithm. In practice, Alpha-Beta variants typically evaluate less nodes than SSS*. A new instance of this framework, MTD($f$), out-performs SSS* and NegaScout, the Alpha-Beta variant of choice by practitioners.

## 1 Introduction

Game playing is one of the classic problems of artificial intelligence. Searching for the best move in a zero-sum game is known as minimax search. The study of minimax search algorithms has created ideas that have proved useful in many search domains. For example, this research extends to single-agent search, such as iterative deepening (IDA*) [11], real-time search (RTA*) [12] and bidirectional search [14]. Although two-agent search has many more application areas, the best-known is undoubtedly game playing, the application that we shall be using in this paper.

Over the last thirty years, most practitioners used depth-first search algorithms based on Alpha-Beta [10] for their game-playing programs. There is an exponential gap in the size of trees built by best-case and worst-case Alpha-Beta. This led to numerous enhancements to the basic algorithm, including iterative deepening, transposition tables, the history heuristic and narrow search windows (see for example [31] for an assessment). Although best-first approaches have been successful in other search domains, minimax search in practice has been almost exclusively based on depth-first strategies. Best-first approaches were more complex and reportedly required more memory, both being serious impediments to their general acceptance. SSS*, a best-first algorithm, will provably never build larger trees than Alpha-Beta and generally builds significantly smaller trees [4, 9, 16, 26, 33]. Despite the potential, the algorithm remains largely ignored in practice.

This paper presents the surprising result that best-first SSS* can be reformulated as a special case of depth-first Alpha-Beta. Consequently, SSS* is now easily implemented in existing Alpha-Beta-based game-playing programs, solving all of the perceived drawbacks of the algorithm. Experiments conducted with three tournament-quality game-playing programs show that in practice SSS* requires as much memory as Alpha-Beta. When given identical memory resources, SSS* does not evaluate significantly less nodes than Alpha-Beta. It is typically out-performed by NegaScout [8, 28, 26], the current depth-first Alpha-Beta variant of choice. In effect the reasons for ignoring SSS* have been eliminated, but the reasons for using it are gone too!

The ideas at the basis of the SSS* reformulation are generalized to create a framework for best-first fixed-depth minimax search that is based on depth-first null-window Alpha-Beta calls. A number of other algorithms in the literature, including DUAL* and C*, are just special cases of this framework. A new instance of this framework, MTD($f$), out-performs all other minimax search algorithms.

In the new framework, SSS* is equivalent to a special case of Alpha-Beta and it is out-performed by other Alpha-Beta variants (both best-first and depth-first). In light of this, we believe that SSS* should now become a footnote in the history of game-tree search.

## 2 Null-Window Search and Memory

In the Alpha-Beta procedure, a node is searched with a search window. It is well-known that the narrower the search window, the more nodes can be cutoff [16]. The narrowest window possible is the null window, where $\alpha = \beta - 1$ (assuming integer-valued leaves). Many people have noted that null-window search has a great potential for creating efficient search algorithms [1, 5, 6, 8, 19, 30]. The widely used NegaScout algorithm derives its superiority over Alpha-Beta from null-window search [8, 20, 26]. A number of algorithms have been proposed that are solely based on null-window search, such as C* [6, 34] and Alpha Bounding [30].

Knuth and Moore have shown that the return value $g$ of an Alpha-Beta search with window $\langle \alpha, \beta \rangle$ can be one of three things [10]:

1. $\alpha < g < \beta$. $g$ is equal to the minimax value $f$ of the game tree $G$.

2. $g \leq \alpha$ (failing low). $g$ is an upper bound on the minimax value $f$ of $G$, or $f \leq g$.

3. $g \geq \beta$ (failing high). $g$ is a lower bound on the minimax value $f$ of $G$, or $f \geq g$.

Knuth and Moore have shown that the essential part of the search tree that proves the minimax value is the *minimal tree* [10]. For a minimax tree of uniform width $w$ and depth $d$, it has $w^{\lfloor d/2 \rfloor} + w^{\lceil d/2 \rceil} - 1$ leaves, or, its size is $O(w^{\lceil d/2 \rceil})$. If Alpha-Beta returns an upper bound, then its value is defined by a max solution tree, in a sense one half of a minimal tree, of size $O(w^{\lceil d/2 \rceil})$. If Alpha-Beta returns a lower bound, then its value is defined by a min solution tree, the other half of a minimal tree, of size $O(w^{\lfloor d/2 \rfloor})$. The theoretical background for these statements can be found in [7, 13, 25, 33].

A single null-window search will never return the true minimax value $f$, but only a bound on it (which may happen to coincide with $f$, but this cannot be inferred from the result of the null-window call). A fail low results in an upper bound, denoted by $f^+$. A fail high returns a lower bound, denoted by $f^-$. Algorithms like C* and Alpha Bounding use multiple null-window calls to generate bounds to home in on the minimax value. A potential problem with these repetitive calls is that they re-expand previously searched nodes. For NegaScout it appears that the gains of the tighter bounds out-weigh the costs of re-expansions, compared to a single wide-window Alpha-Beta call [21].

An idea to prevent the re-expansion of previously searched nodes is to store them in memory. It is often said that since minimax trees are of exponential size, this approach is infeasible since it needs exponentially growing amounts of memory [9, 18, 29]. For game-playing programs an obvious choice is to use a transposition table for storage [31]. Originally the transposition table was introduced to prevent the search of transpositions in the search space. A transposition is a node with more than one parent, a position that can be reached via several move sequences. Today, transposition tables are often used in combination with iterative deepening [16]. The main benefit of this combination is to improve the quality of move ordering [15].

Storing information from previous searches is another use of the transposition table. The only difference with preventing the search of transpositions is that now the table entries are nodes from a previous search pass; there is no difference as far as the table itself is concerned. The basic idea is that it is possible to store the search tree in the transposition table. Some background explaining in greater detail why the transposition table is a suitable structure for storing search or solution trees, and why it gives correct results in algorithms doing repetitive null-window searches, can be found in [7, 25].

Since recursive null-window Alpha-Beta calls return only bounds, storing the previous search results comes down to storing a max or a min solution tree. We have shown in [25] that although the search information that must be stored is indeed of exponential size, it is much less than what is often assumed. For the search depths typically reached by tournament-quality game-playing programs, the search information fits comfortably in today's memories. Projections of tomorrow's search depths and memory sizes show that this situation will persist in the foreseeable future.

## 3 A Framework for Best-First Search

The concept of null-window Alpha-Beta search was introduced by Pearl with his proof-procedure Test [20]. Since we use memory to store intermediate search information, we have named our framework Memory-enhanced Test, or MT. As proof procedure we use a standard null-window Alpha-Beta procedure for use with transposition tables. The code can be found in figure 1.

So far, we have discussed the following two mechanisms to be used in building efficient algorithms: (1) null-window searches cutoff more nodes than wide search windows, and (2) transposition tables can be used to glue multiple Alpha-Beta passes together. We can use these building blocks to construct a number of different algorithms. An option is to construct drivers that repeatedly call Alpha-Beta at the root of the game tree. Three of these drivers are shown in the figures 2 and 3. The drivers differ in the way the null window is chosen (denoted by $\gamma$ in the figures).

### 3.1 SSS*

The driver to the left in figure 2 constructs an algorithm that starts with an upper bound of $+\infty$. From Alpha-Beta's postcondition we see that this call will fail low, yielding an upper bound. By feeding this upper bound $f^+$ again to a null-window Alpha-Beta call, we will get a sequence of fail lows. In the end, if $g = \gamma$, we will have a fail high with $g = f^- = \gamma = f^+$, which means the

```
/* Transposition table (TT) enhanced Alpha-Beta */
function Alpha-Beta(n, α, β) → g;
    /* Check if position is in TT and has been searched to sufficient depth */
    if retrieve(n) = found then
        if n.f⁺ ≤ α or n.f⁺ = n.f⁻ then return n.f⁺;
        if n.f⁻ ≥ β then return n.f⁻;
    /* Reached the maximum search depth */
    if n = leaf then
        n.f⁻ := n.f⁺ := g := eval(n);
    else
        g := −∞; a := α;
        c := firstchild(n);
        /* Search until a cutoff occurs or all children have been considered */
        while g < β and c ≠ ⊥ do
            g := max(g, −Alpha-Beta(c, −β, −a));
            a := max(a, g);
            c := nextbrother(c);
        /* Save in transposition table */
        if g ≤ α then n.f⁺ := g;
        if α < g < β then n.f⁺ := n.f⁻ := g;
        if g ≥ β then n.f⁻ := g;
    store(n);
    return g;
```

Figure 1: Alpha-Beta for Use with Transposition Tables

minimax value $f$ has been found. This driver expands the same leaf nodes in the same order as Stockman's SSS* [33]. (A full proof of this claim can be found in [22] and an outline of the proof in [25].) In this sense, we have constructed an equivalent formulation of SSS*, constructing a best-first algorithm using depth-first, memory-enhanced, search. The reformulation is called AB-SSS*.

Many researchers have conjectured that best-first algorithms such as SSS* would need too much memory to be practical alternatives for depth-first algorithms like Alpha-Beta. The literature cites three main drawbacks of SSS*: it is hard to understand, it performs operations on a sorted list that are slow, and it uses too much memory to be practical [9, 17, 18, 26, 29, 33]. The new formulation has a number of practical advantages over the old Stockman formulation. The biggest advantage is that this formulation is readily implementable in a regular Alpha-Beta-based game-playing program. This enables us to easily test the performance of SSS* These tests confirm that SSS* does *not* need too much memory [25].

We think that our reformulation as a sequence of null-window Alpha-Beta calls is easy to understand. SSS*'s slow OPEN list operations are traded in for hash table lookups that are as fast as for Alpha-Beta [16], and the experiments show that AB-SSS* does not need too much memory. We conclude that the drawbacks of SSS* are solved in the new formulation.

```
function AB-SSS*(n) → f;                function AB-DUAL*(n) → f;
    g := +∞;                                g := −∞;
    repeat                                  repeat
        γ := g;                                 γ := g;
        g := Alpha-Beta(n, γ − 1, γ);           g := Alpha-Beta(n, γ, γ + 1);
    until g = γ;                            until g = γ;
    return g;                               return g;
```

Figure 2: SSS* and DUAL* as a Sequence of Memory-enhanced Alpha-Beta Searches

```
function MTD(n, f) → f;
    g := f;
    f⁺ := +∞; f⁻ := −∞;
    repeat
        if g = f⁻ then γ := g + 1 else γ := g;
        g := Alpha-Beta(n, γ − 1, γ);
        if g < γ then f⁺ := g else f⁻ := g;
    until f⁻ = f⁺;
    return g;
```

Figure 3: MTD($f$), a Better Sequence of Alpha-Beta Searches

## 3.2 DUAL*

A dual version of SSS*, aptly named DUAL*, can be created by inverting SSS*'s operations: use an ascendingly sorted list instead of descending, swap max and min operations, and start at −∞ instead of +∞ [17, 26]. The power of the framework is demonstrated by the reformulation called AB-DUAL* in figure 2. The only difference with AB-SSS* is the initialization of the bound to −∞, and a change in the way Alpha-Beta is called. This reformulation focuses attention on one item only: the bound starts at the bottom of the scale, implying that the only fundamental difference between SSS* and DUAL* is that upper bounds are replaced by lower bounds (which implies that the max solution tree that is refined by AB-SSS* has become a min solution tree in AB-DUAL*). All other differences are apparently insubstantial, since nothing else has to be changed.

## 3.3 Other Options for the choice of Start Value

AB-SSS* starts the sequence of Alpha-Beta searches at +∞, the high end of the scale. AB-DUAL* starts at −∞, the low end of the scale. An intuitively appealing option is to choose another start value, closer to the expected outcome. One option is to keep bisecting the interval between the upper and lower bound, to reduce the number of Alpha-Beta calls. This idea is used in C* [6, 23, 34]. Another idea is to use a heuristic guess as the start value. In an iterative deepening framework it is natural to use the score from the previous iteration for this purpose, since it is expected to be a close approximation of the score for the current depth. We have called this driver MTD($f$) and its pseudo code is shown in figure 3. The first call acts to decide which way the search will go. If it is a fail high, MTD($f$) will behave like AB-DUAL*, and keep increasing the lower bound returned by Alpha-Beta. If the first call fails low, MTD($f$) will, like AB-SSS*, decrease the upper bound until the minimax value is reached. AB-SSS* starts off optimistic, AB-DUAL* starts off pessimistic, and MTD($f$) starts off in the middle, possibly realistic.

One of the drawbacks of AB-SSS* and AB-DUAL* is the potentially high number of calls to Alpha-Beta needed until the search converges to the minimax value. Most of the Alpha-Beta calls make small improvements to the bound. By starting closer to the minimax value, many intermediate Alpha-Beta calls are skipped. MTD($f$) takes one big leap to come close to the minimax value, dramatically reducing the number of intermediate Alpha-Beta calls. The lower number of calls has the advantage that MTD($f$) performs relatively better in constrained memory than SSS*, since there are fewer re-expansions. Measurements confirm that Alpha-Beta typically gets called 3 to 6 times in MTD($f$). In contrast, the AB-SSS* and AB-DUAL* results are poor compared to NegaScout when all nodes in the search tree are considered. Each of these algorithms performs dozens and often even hundreds of Alpha-Beta searches. The wider the range of leaf values, the smaller the steps with which they converge, and the more re-searches they need.

## 4 Performance

To assess the performance of the proposed algorithms in practice, a series of experiments was conducted. We present data for the comparison of Alpha-Beta, NegaScout, AB-SSS*, AB-DUAL*, and MTD($f$).

### 4.1 Experiment Design

We will assess the performance of the algorithms by counting leaf nodes and total nodes. For two algorithms we also provide data for execution time. This metric may vary considerably for different programs. It is nevertheless included, to give evidence of the potential of MTD($f$).

We have tried to come as close to real-life applications of our algorithms as possible by conducting the experiments with three tournament-quality game-playing programs, Phoenix [30] for chess, Keyano [3] for Othello and Chinook [32] for checkers. Chess has a branching factor of about 35, Othello of about 10 and checkers of about 3. Thus we cover the range from a wide to a narrow branching factor. This paper presents results for chess; the results for the other games (which are similar and confirm the chess results) can be found in [25]. All algorithms used iterative deepening. They are repeatedly called with successively deeper search depths. All three algorithms use a standard transposition table with a maximum of $2^{21}$ entries; tests showing that the solution trees could comfortably fit in tables of this size, without any risk of noise due to collisions [25]. For our experiments we used the original program author's transposition table data structures and code without modification. At an interior node, the move suggested by the transposition table is always searched first (if known), and the remaining moves are ordered before being searched. Phoenix uses dynamic ordering based on the history heuristic [31].

Conventional test sets in the literature proved to be inadequate to model real-life conditions. Positions in test sets are usually selected to test a particular characteristic or property of the game (such as tactical combinations in chess) and are not necessarily indicative of typical game conditions. For our experiments, the algorithms were tested using a set of 20 positions that corresponded to move sequences from tournament games. By selecting move sequences rather than isolated positions, we are attempting to create a test set that is representative of real game search properties (including positions with obvious moves, hard moves, positional moves, tactical moves, different game phases, etc.). A number of test runs was performed on a bigger test set and to a higher search depth to check that the 20 positions did not contain anomalies.

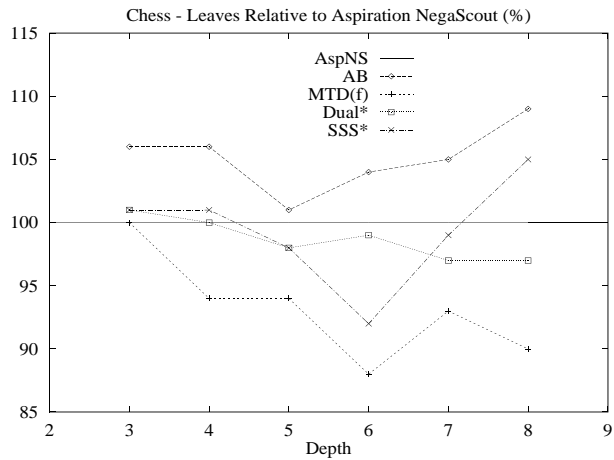

Figure 4: Comparing Leaf Node Count

Many papers in the literature use Alpha-Beta as the base-line for comparing the performance of other algorithms (for example, [5, 15]). The implication is that this is the standard data point which everyone is trying to beat. However, game-playing programs have evolved beyond simple Alpha-Beta algorithms. Most use Alpha-Beta enhanced with null-window search (NegaScout), iterative deepening, transposition tables, move ordering and an initial aspiration window. Since this is the typical search algorithm used in high-performance programs (such as Phoenix), it seems more reasonable to use this as our base-line. The worse the base-line comparison algorithm chosen, the better other algorithms appear to be. By choosing NegaScout enhanced with aspiration searching [2] (Aspiration NegaScout) as our performance metric, and giving it a transposition table big enough to contain all re-search information, we are emphasizing that it is possible to do better than the "best" methods currently practiced and that, contrary to published simulation results, some methods—notably SSS*—will turn out to be inferior.

Since we implemented the algorithms (like AB-SSS* and AB-DUAL*) using Alpha-Beta we were able to compare a number of algorithms that were previously seen as very different. By using Alpha-Beta as a common procedure, every algorithm benefited from the same enhancements concerning iterative deepening, transposition tables and move ordering code. To our knowledge this is the first comparison of depth-first and best-first minimax search algorithms where all the algorithms are given identical resources.

### 4.2 Results

Figure 4 shows the performance of Phoenix using the number of leaf evaluations (NBP or Number of Bottom Positions) as the performance metric. Figure 5 shows the performance of the algorithms using the number of nodes in the search tree (interior and leaf, including nodes that caused transposition cutoffs) as the metric. The graphs show the cumulative number of nodes over all previous iterations for a certain depth (which is realistic since iterative deepening is used) relative to Aspiration NegaScout. Note the different vertical scales.

#### 4.2.1 Aspiration NegaScout and MTD($f$)

The results show that Aspiration NegaScout is better than Alpha-Beta. This result is consistent with [31] which showed Aspiration NegaScout to be a small improvement over Alpha-Beta

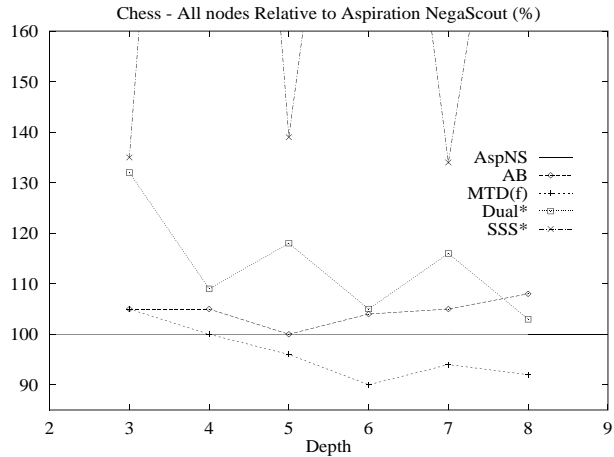

Figure 5: Comparing Total Node Count

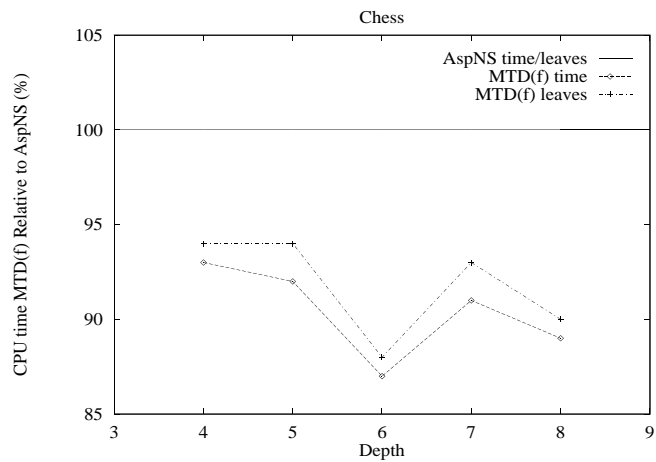

Figure 6: Comparing Execution Time

when transposition tables and iterative deepening were used.

MTD($f$) is a best-first algorithm that consists solely of null-window searches. In each pass, the previous one is used to guide the search towards selecting the best node. The majority of the searches in NegaScout is also performed with a null window. An important difference is with which value this null-window search is performed. NegaScout derives it from the tree itself, whereas MTD($f$) relies for the first guess on information from outside the tree. (In our experiments the minimax value from a previous iterative deepening iteration was used for this purpose.)

The best results are from MTD($f$), although the current algorithm of choice by the game programming community, Aspiration NegaScout, performs very well too. The averaged MTD($f$) leaf node counts are consistently better than for Aspiration NegaScout, averaging around a 10% improvement for Phoenix. More surprisingly is that MTD($f$) outperforms Aspiration NegaScout on the total node measure as well. Since each iteration requires repeated calls to Alpha-Beta (at least two and possibly many more), one might expect MTD($f$) to perform badly by this measure because of the repeated traversals of the tree. This suggests that MTD($f$), on average, is calling Alpha-Beta close to the minimum number of times. A deeper analysis of MTD($f$) can be found in [25].

### 4.2.2 SSS* and DUAL*

Contrary to many simulations [4, 9, 17, 18, 26, 27, 29, 33], our results show that the difference in the number of leaves expanded by SSS* and Alpha-Beta is relatively small. Since game-playing programs use many search enhancements, the benefits of a best-first search are greatly reduced. We conclude that in practice, AB-SSS* is a small improvement on Alpha-Beta for leaf nodes only (depending on the branching factor). Claims that SSS* and DUAL* evaluate significantly fewer leaf nodes than Alpha-Beta are based on simplifying assumptions that have little relation with what is used in practice. Aspiration NegaScout regularly out-performs SSS* on leaf count and greatly out-performs it on total nodes.

### 4.3 Execution time

The bottom line for practitioners is execution time. We only show execution time graphs for iterative deepening (ID) MTD($f$) and ID Aspiration NegaScout (figure 6). Comparing results for the same machines we found that MTD($f$) is consistently the fastest algorithm. The run shown is a typical example run on a Sun SPARC. We did experience different timings when running on different machines. It may well be that cache size plays an important role, and that tuning of the program can have a considerable impact.

In our experiments we found that for Phoenix MTD($f$) was about 9–13% faster in execution time than Aspiration NegaScout. For other programs and other machines these results will obviously differ, depending in part on the quality of the score of the previous iteration, and on the test positions used. Also, since the tested algorithms perform quite close together, the relative differences are quite sensitive to variations in input parameters. In generalizing these results, one should keep this sensitivity in mind. Using these numbers as absolute predictors for other situations would not do justice to the complexities of real life game trees. The experimental data is better suited to provide insight on, or guide and verify hypotheses about these complexities.

### 4.4 Artificial versus Real Trees

Previously, performance assessments of SSS* and Alpha-Beta have mainly been based on simulations [4, 9, 17, 18, 26, 27, 29, 33]. They typically found SSS* to out-perform Alpha-Beta significantly. Our experiments were performed with practical programs, that enhance the basic minimax algorithms with techniques like iterative deepening and transposition tables. The result is that in practice SSS* does not out-perform Alpha-Beta significantly. It is often out-performed by depth-first Alpha-Beta variants such as NegaScout.

Simulations are usually performed when it is too difficult or too expensive to construct the proper experimental environment. For game-tree searching, the case for simulations is weak. There is no need to do simulations when there are quality game-playing programs available for obtaining actual data. Further, simulation parameters can be incorrect, resulting in large errors in the results that lead to misleading conclusions.

## 5 Conclusions

Null-window search, enhanced with storage, can be used to construct best-first minimax algorithms. For storage a conventional transposition table can be used. The null-window calls generate a sequence of bounds on the minimax value. The storage contains the part of the search tree that establishes these bounds, to be refined in subsequent passes.

A framework has been presented for algorithms that generate sequences of bounds in different ways. Interestingly, certain instances of this framework expand the same leaf nodes in the same order as SSS* and DUAL*. These algorithms, called AB-SSS* and AB-DUAL*, solve the perceived problems of SSS* and DUAL*. They are much simpler and practical, consisting of a single loop of Alpha-Beta calls.

We used tournament game-playing programs for our tests. Using the Alpha-Beta-based framework, both depth-first and best-first algorithms are given the same storage, in contrast to previous comparisons. The results confirm that AB-SSS* and AB-DUAL* are practical algorithms, contradicting previous publications [9, 17, 29].

We tested instances of the framework against the depth-first algorithm implemented in our programs, Aspiration NegaScout, representing the current choice of the game programming community. One instance, MTD($f$), out-performs NegaScout on leaf node count, total node count and execution time, by a wider margin than NegaScout's gain over Alpha-Beta. The results reported in this paper are for chess, a game with a relatively high branching factor. We have conducted experiments for two other games as well. The results for checkers and Othello, games with a narrow and medium branching factor, confirm these results [25].

Interestingly, all the tested algorithms perform relatively close together, much closer than previous simulation results have indicated. We conclude that, a) artificially-generated game trees often do not capture all the essential aspects of real trees, and b) often more gains are obtained from the so-called Alpha-Beta enhancements, than from the underlying algorithms.

In the past, much research effort has been devoted to understanding how SSS* works, and finding out what the pros and cons of SSS*'s best-first approach are for minimax search. In the new framework, SSS* is equivalent to a special case of Alpha-Beta and it is out-performed by other Alpha-Beta variants (both best-first and depth-first). In light of this, we believe that SSS* should now become a footnote in the history of game-tree search.

Having seen these results for minimax search, it is an interesting question to find out whether formulating best-first search using depth-first procedures is possible in single-agent search as well.

## Acknowledgements

Some of these results were originally published in [24].

This work has benefited from discussions with Mark Brockington, Yngvi Bjornsson and Andreas Junghanns. The support of Jaap van den Herik, and the financial support of the Netherlands Organization for Scientific Research (NWO), the Tinbergen Institute, the Natural Sciences and Engineering Research Council of Canada (NSERC grant OGP-5183) and the University of Alberta Central Research Fund are gratefully acknowledged.

## References

[1] L. Victor Allis, Maarten van der Meulen, and H. Jaap van den Herik. Proof-number search. *Artificial Intelligence*, 66:91–124, March 1994.

[2] Gérard M. Baudet. *The Design and Analysis of Algorithms for Asynchronous Multiprocessors.* PhD thesis, Carnegie Mellon University, Pittsburgh, PA, 1978.

[3] Mark Brockington. *Asynchronous Parallel Game-Tree Search.* PhD thesis, proposal, Department of Computing Science, University of Alberta, Edmonton, Canada, 1994.

[4] Murray Campbell. Algorithms for the parallel search of game trees. Master's thesis, Department of Computing Science, University of Alberta, Canada, August 1981.

[5] Murray S. Campbell and T. Anthony Marsland. A comparison of minimax tree search algorithms. *Artificial Intelligence*, 20:347–367, 1983.

[6] K. Coplan. A special-purpose machine for an improved search algorithm for deep chess combinations. In M.R.B. Clarke, editor, *Advances in Computer Chess 3, April 1981*, pages 25–43. Pergamon Press, Oxford, 1982.

[7] Arie de Bruin, Wim Pijls, and Aske Plaat. Solution trees as a basis for game-tree search. *ICCA Journal*, 17(4):207–219, December 1994.

[8] John P. Fishburn. *Analysis of Speedup in Distributed Algorithms.* PhD thesis, University of Wisconsin, Madison, 1981.

[9] Hermann Kaindl, Reza Shams, and Helmut Horacek. Minimax search algorithms with and without aspiration windows. *IEEE Transactions on Pattern Analysis and Machine Intelligence*, 13(12):1225–1235, December 1991.

[10] Donald E. Knuth and Ronald W. Moore. An analysis of alpha-beta pruning. *Artificial Intelligence*, 6(4):293–326, 1975.

[11] Richard E. Korf. Iterative deepening: An optimal admissible tree search. *Artificial Intelligence*, 27:97–109, 1985.

[12] Richard E. Korf. Real-time heuristic search. *Artificial Intelligence*, 42:189–211, 1990.

[13] Vipin Kumar and Laveen N. Kanal. A general branch and bound formulation for and/or graph and game tree search. In *Search in Artificial Intelligence.* Springer Verlag, 1988.

[14] J.B.H. Kwa. BS*: An admissible bidirectional staged heuristic search algorithm. *Artificial Intelligence*, 38:95–109, February 1989.

[15] T. Anthony Marsland. A review of game-tree pruning. *ICCA Journal*, 9(1):3–19, March 1986.

[16] T. Anthony Marsland and Murray S. Campbell. Parallel search of strongly ordered game trees. *Computing Surveys*, 14(4):533–551, December 1982.

[17] T. Anthony Marsland, Alexander Reinefeld, and Jonathan Schaeffer. Low overhead alternatives to SSS*. *Artificial Intelligence*, 31:185–199, 1987.

[18] Agata Muszycka and Rajjan Shinghal. An empirical comparison of pruning strategies in game trees. *IEEE Transactions on Systems, Man and Cybernetics*, 15(3):389–399, May/June 1985.

[19] Judea Pearl. Asymptotical properties of minimax trees and game searching procedures. *Artificial Intelligence*, 14(2):113–138, 1980.

[20] Judea Pearl. The solution for the branching factor of the alpha-beta pruning algorithm and its optimality. *Communications of the ACM*, 25(8):559–564, August 1982.

[21] Judea Pearl. *Heuristics – Intelligent Search Strategies for Computer Problem Solving.* Addison-Wesley Publishing Co., Reading, MA, 1984.

[22] Wim Pijls, Arie de Bruin, and Aske Plaat. Solution trees as a unifying concept for game tree algorithms. Technical Report EUR-CS-95-01, Erasmus University, Department of Computer Science, P.O. Box 1738, 3000 DR Rotterdam, The Netherlands, April 1995.

[23] Aske Plaat, Jonathan Schaeffer, Wim Pijls, and Arie de Bruin. A new paradigm for minimax search. Technical Report TR-CS-94-18, Department of Computing Science, University of Alberta, Edmonton, AB, Canada, December 1994.

[24] Aske Plaat, Jonathan Schaeffer, Wim Pijls, and Arie de Bruin. Best-first fixed-depth game-tree search in practice. In *Proceedings of the International Joint Conference on Artificial Intelligence (IJCAI-95)*, volume 1, pages 273–279, August 1995.

[25] Aske Plaat, Jonathan Schaeffer, Wim Pijls, and Arie de Bruin. A minimax algorithm better than Alpha-Beta? no and yes. Technical Report 95-15, University of Alberta, Department of Computing Science, Edmonton, AB, Canada T6G 2H1, May 1995.

[26] Alexander Reinefeld. *Spielbaum Suchverfahren*. Informatik-Fachberichte 200. Springer Verlag, 1989.

[27] Alexander Reinefeld and Peter Ridinger. Time-efficient state space search. *Artificial Intelligence*, 71(2):397–408, 1994.

[28] Alexander Reinefeld, Jonathan Schaeffer, and T. Anthony Marsland. Information acquisition in minimal window search. In *Proceeding of the International Joint Conference on Artificial Intelligence (IJCAI-85)*, volume 2, pages 1040–1043, 1985.

[29] Igor Roizen and Judea Pearl. A minimax algorithm better than alpha-beta? Yes and no. *Artificial Intelligence*, 21:199–230, 1983.

[30] Jonathan Schaeffer. *Experiments in Search and Knowledge.* PhD thesis, Department of Computing Science, University of Waterloo, Canada, 1986. Available as University of Alberta technical report TR86-12.

[31] Jonathan Schaeffer. The history heuristic and alpha-beta search enhancements in practice. *IEEE Transactions on Pattern Analysis and Machine Intelligence*, 11(1):1203–1212, November 1989.

[32] Jonathan Schaeffer, Joseph Culberson, Norman Treloar, Brent Knight, Paul Lu, and Duane Szafron. A world championship caliber checkers program. *Artificial Intelligence*, 53(2-3):273–290, 1992.

[33] George C. Stockman. A minimax algorithm better than alpha-beta? *Artificial Intelligence*, 12(2):179–196, 1979.

[34] Jean-Christophe Weill. The NegaC* search. *ICCA Journal*, 15(1):3–7, March 1992.